\documentclass[twoside,11pt]{article}

\usepackage{jmlr2e}
\usepackage{amsmath}
\usepackage{longtable}
\usepackage{xcolor}
\usepackage{listings}
\usepackage{color}
\usepackage{soul}
\usepackage{multirow}
\usepackage{booktabs}
\usepackage{enumitem,kantlipsum}
\usepackage{amsfonts}
\usepackage{balance}
\usepackage{lipsum}
\usepackage{booktabs,makecell}
\usepackage{arydshln}
\usepackage{placeins}

\ShortHeadings{\textsc{NeurIPS2022 NL4Opt} Competition}{Ramamonjison et al.}
\firstpageno{1}

\begin{document}

\title{\textsc{NL4Opt Competition}: Formulating Optimization Problems \\ Based on Their Natural Language Descriptions}

\author{\name Rindranirina Ramamonjison\textsuperscript{1} \email rindranirina.ramamonjison@huawei.com
       \AND
       \name Timothy T. Yu\textsuperscript{1} \email timothyt.yu@huawei.com 
       \AND
       \name Raymond Li\textsuperscript{2} \email raymondl@cs.ubc.ca 
       \AND
       \name Haley Li\textsuperscript{1,2} \email haleyli@cs.ubc.ca 
       \AND
       \name Giuseppe Carenini\textsuperscript{2} \email carenini@cs.ubc.ca
       \AND
       \name Bissan Ghaddar\textsuperscript{3} \email bghaddar@ivey.ca 
       \AND
       \name Shiqi He\textsuperscript{1,2} \email shiqihe@cs.ubc.ca 
       \AND
       \name Mahdi Mostajabdaveh\textsuperscript{1} \email mahdi.mostajabdaveh1@huawei.com 
       \AND
       \name Amin Banitalebi-Dehkordi\textsuperscript{1} \email amin.banitalebi@gmail.com 
       \AND
       \name Zirui Zhou\textsuperscript{1} \email zirui.zhou@huawei.com 
       \AND
       \name Yong Zhang\textsuperscript{1} \email yong.zhang3@huawei.com \\ \\
       \addr \textsuperscript{1} Huawei Technologies Canada, Burnaby, Canada\\
       \addr \textsuperscript{2} University of British Columbia, Vancouver, Canada\\
       \addr \textsuperscript{3} Ivey Business School, London, Canada
       }

\editor{Marco Ciccone, Gustavo Stolovitzky, Jake Albrecht}

\maketitle

\begin{abstract}
The \textsc{Natural Language for Optimization (NL4Opt) Competition} was created to investigate methods of extracting the meaning and formulation of an optimization problem based on its text description. Specifically, the goal of the competition is to increase the accessibility and usability of optimization solvers by allowing non-experts to interface with them using natural language. We separate this challenging goal into two sub-tasks: (1) recognize and label the semantic entities that correspond to the components of the optimization problem; (2) generate a meaning representation (i.e. a logical form) of the problem from its detected problem entities. The first task aims to reduce ambiguity by detecting and tagging the entities of the optimization problems. The second task creates an intermediate representation of the linear programming (LP) problem that is converted into a format that can be used by commercial solvers. In this report, we present the LP word problem dataset and shared tasks for the NeurIPS 2022 competition. Furthermore, we investigate and compare the performance of the ChatGPT large language model against the winning solutions. Through this competition, we hope to bring interest towards the development of novel machine learning applications and datasets for optimization modeling. \end{abstract}


\section{Introduction}

Operations research (OR) tools can be leveraged to model and solve many real-world decision-making problems analytically and efficiently. OR is a field of applied mathematics that has been proven beneficial in many applications such as supply chain management \citep{supplychain}, production planning \citep{productionplanning}, bike-share ridership and efficiency in urban cities \citep{bikeshare, bicycleGuangzhou}, managing wastewater collection and treatment systems \citep{wastewater}, and finding a revenue-maximizing pricing strategy for businesses \citep{revenue}. Different types of optimization problems can be solved using standard optimization algorithms such as the simplex \citep{simplex} or interior-point method \citep{karmarkar}. However, modeling real-world problems into proper formulations as input to optimization solvers is still an iterative and strenuous process. First, the problem must be described by the stakeholder in the language of a domain expert. Then, an OR expert must extract the decision variables, objective, and the constraints from the description. Finally, the problem must be re-written in an algebraic modeling language that solvers can interpret.

Through the \textsc{NL4Opt Competition}, we investigate the feasibility of learning-based natural language interfaces for optimization solvers. To do so, we explored the practicality of partially automating the formulation of optimization problems. In particular, semantic parsing is a general task for extracting machine-interpretable meaning representations from natural language utterances. They have been well-studied for designing NLP systems that interact with database systems \citep{zhong2017seq2sql-custom, gan-etal-2020-review}, Unix machines \citep{lin-etal-2018-nl2bash}, knowledge base systems \citep{berant-liang-2014-semantic, dong-lapata-2016-language} or dialog systems \citep{guo-2018-custom}. However, extracting the formulation of optimization problems is still an under-explored problem. Meanwhile, solving math word problems with NLP has seen sustained research activity \citep{koncel-kedziorski-etal-2016-mawps, hopkins-etal-2019-semeval, miao-etal-2020-diverse, patel-etal-2021-nlp} with researchers focused on finding the correct answers to elementary algebraic and arithmetic problems. In contrast, rather than exploring methods of producing the solution to the problem, we focus on converting optimization problems into a form that can be passed to commercial optimization solvers to efficiently find optimal solutions. 

Lately, a few related challenges have been created for analyzing scientific texts. For instance, \cite{harper-etal-2021-semeval} proposed the MeasEval challenge focused on extracting counts and measurements from clinical documents and finding the attributes of those quantities. Another popular challenge was MultiCoNER \citep{multiconer} which focused on detecting semantically ambiguous and complex entities from documents written in 11 languages spanning 13 tracks. The \textsc{NL4Opt Competition} expands this task by not only detecting complex entities from optimization problems but also generating the equivalent mathematical formulation. 
\section{The \textsc{NL4Opt Competition}}

\begin{figure*}
\centering\includegraphics[width=0.6\columnwidth]{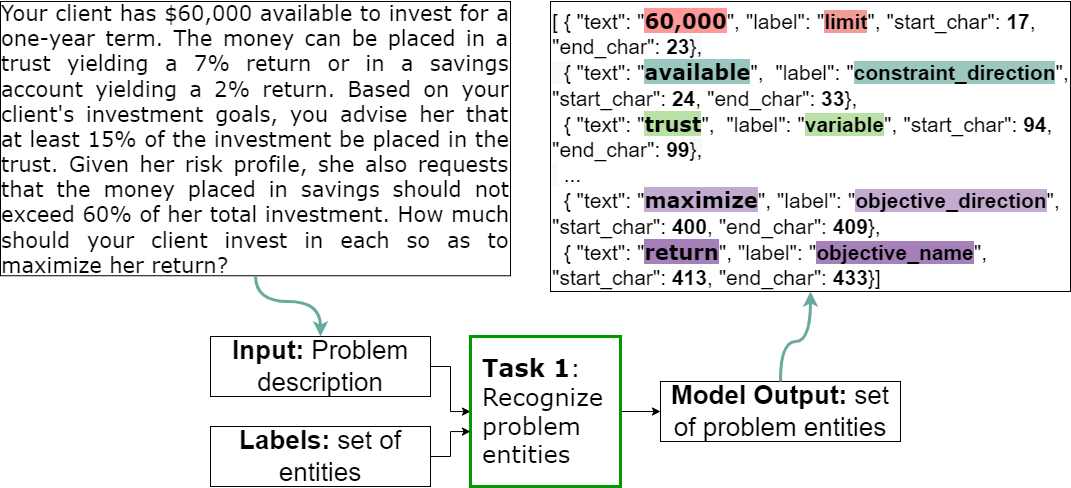}
\caption{Description of dataset for sub-task 1: ``recognizing the problem entities"}
\label{fig:subtask-1}
\end{figure*}

The \textsc{NL4Opt Competition} explores the design of natural language interfaces for optimization solvers. The results of this competition forward the accessibility and usability of these solvers and allow non-experts to solve important problems from various industries. Specifically, we used this competition to explore methods of converting a natural language description of an optimization problem into a mathematical formulation. This goal was separated into two inter-related sub-tasks:
\begin{enumerate}
    \item Recognition of optimization problem entities,
    \item Generation of problem formulation.
\end{enumerate}

The first sub-task was to recognize optimization model entity types (i.e., constraint direction, constraint limit, objective direction, objective name, parameter, variable) from the problem description. In the first sub-task, the goal was to detect text spans from the problem description that represent semantic entities of the optimization problem and to tag them according to the listed entity types. This sub-task aimed to reduce the ambiguity by identifying important components of the optimization model. An illustration of sub-task 1 is provided in Figure \ref{fig:subtask-1}. 

\begin{figure*}
\centering\includegraphics[width=0.8\columnwidth]{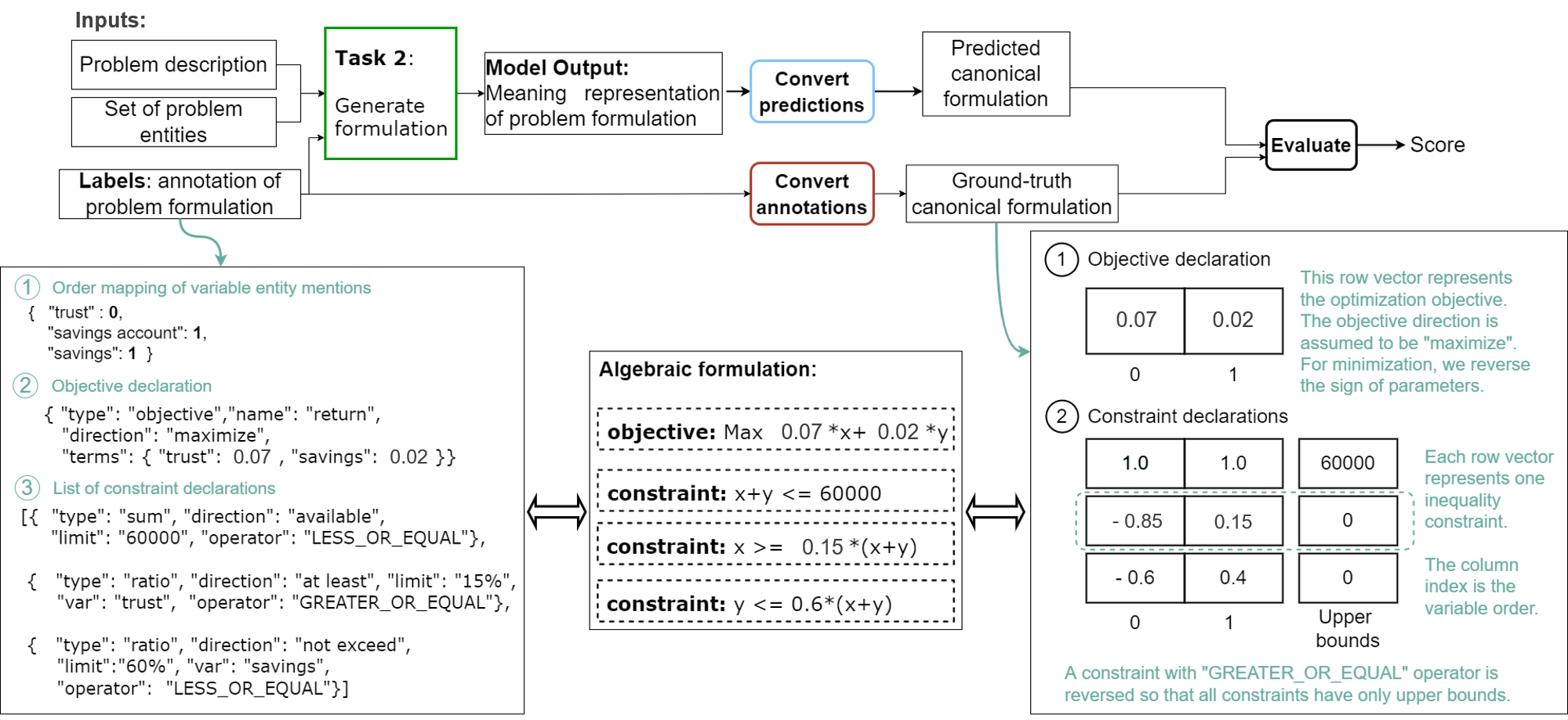}
\caption{Dataset annotation and evaluation protocol for sub-task 2: "generating the problem formulation"}
\label{fig:subtask-2}
\end{figure*}

The second sub-task was to generate a precise meaning representation of the optimization formulation. This sub-task was simplified using the ground truth information of the problem entities from the first sub-task. An illustration of sub-task 2 is provided in Figure \ref{fig:subtask-2}. \\

\noindent The proposed sub-tasks are characterized by the following challenges: 

\begin{enumerate}
\item \textbf{Unstructured multi-sentence input}. 
An optimization description is the input document that describes the decision variables, objective, and a set of constraints. In addition, the structure of the input varies depending on the structure of the optimization problem and the linguistic style. Thus, the multi-sentence input exhibits a high level of compositionality and ambiguity due to the variability of the linguistic patterns, of the problem domains, and of the problem structures. 
\item \textbf{Mismatched inputs and outputs}. The contextual information from the input description is abstracted away in the target formulation. Therefore, the absence of contextual clues in the output makes it difficult to align the input-output pair. Thus, the meaning representation of the problem formulation (i.e. the output of generation model) is important as it bridges the problem description and the mathematical formulation. In fact, semantically equivalent representations may have syntactically different forms and can lead to different performance \citep{guo-etal-2020-benchmarking-meaning}. 
\item \textbf{Low-resource learning constraint}. Specialized knowledge is required to create a dataset thereby drastically increasing the cost of dataset creation. The design of machine learning models for this task is challenging as they must learn from a small number of expert-annotated examples.
\item \textbf{Domain-agnostic parsing}. Finally, OR tools are applied to disparate problem domains (e.g. forestry, transportation or medicine) \citep{williams2013model}. As a result, the learning-based solution must generalize not only to new problem instances but also to new application domains. 
\end{enumerate}

\subsection{Evaluation}

\paragraph{NER sub-task:} we evaluated the models based on their achieved micro-averaged F1 score given by:
\vspace{-2pt}
\begin{equation}
\text{{F1}}=\frac{2\times P\times R}{P+R},
\end{equation}
where $P$ and $R$ are the average precision and average recall further averaged over all entity types, respectively.

\paragraph{Generation sub-task:} we used an application-specific metric since the task was motivated by the need to precisely formulate the optimization problem. We have benchmarked the models based on the declaration-level mapping accuracy given by:
\vspace{-2pt}
\begin{equation}\label{equ:accuracy}
\text{Acc} = 1 - \frac{\sum_{i=1}^N \min\left\{\text{FP}_i + \text{FN}_i, D_i\right\}}{\sum_{i=1}^N D_i},
\end{equation}
where $N$ is the number of linear programming word problems (LPWPs) in the test set. For each LPWP $i$, $D_i$ is the number of ground-truth declarations. The false positive $\text{FP}_{i}$ is the number of non-matched predicted declarations whereas the false negative $\text{FN}_{i}$ denotes the number of ground-truth declarations without a match. To clarify the evaluation protocol, we have emphasized that the canonical representation, as described in Figure \ref{fig:subtask-2}, would be used to compare the ground-truth and predicted formulations.

\subsection{Competition statistics}
Over 150 teams registered for this competition combining for a total of more than 300 valid independent submissions. The demographics of registered participants were affiliated as follows: 30\% post-secondary, 30\% independent, 30\% unspecified, and 10\% industry.  There were 19 teams with valid submissions to sub-task 1 and 9 to sub-task 2. These teams reported the following affiliations: 60\% industry, 25\% post-secondary, and 15\% independent.

\subsection{Additional Competition Details}

All the details and relevant information of the competition are made accessible at the competition website (\href{https://nl4opt.github.io/}{\textcolor{blue}{https://nl4opt.github.io/}}). This website contains the rules, a FAQ/Tutorial section, access to the starter kit, and final results of the competition. We have released the test set and encourage all new and returning participants to leverage this competition as a benchmark for new methods. We especially welcome those that are interested in tackling the challenges listed above (i.e., unstructured input, misaligned input-output pair, low-resource learning constraint, generalizability).
\section{The \textsc{NL4Opt} Dataset}

\paragraph{Dataset description.} A total of 1101 annotated LPWPs from 6 different domains were created for the \textsc{NL4Opt Competition}. We separated the dataset into 713, 99, and 289 samples for training, development, and testing, respectively. The data samples were distributed identically for both sub-tasks. It is important to evaluate submissions for generalizability towards unseen problem domains. Therefore, we include LPWPs from the three similar (source) domains of sales, advertising, and investment in both training, development and test splits. However, problems from three other (target) domains (production, transportation, and sciences) have been reserved for the development and test splits. To ensure that the development set was never used for training, we reviewed the final submitted code and re-trained all submissions prior to announcing the winning teams. Table \ref{tab:data-distribution} presents the number of samples and the ratio between the source and target domains for the three splits of data. An example of data and its annotations for the two sub-tasks is illustrated in Figures \ref{fig:subtask-1} and \ref{fig:subtask-2}. 

\begin{table}[ht]
\centering
\caption{\label{tab:data-distribution}Data Distribution.}
\begin{tabular}[t]{lcc}
\toprule
Split & \#/samples & source:target\\
\midrule
Train&713&1:0\\
Dev&99&1:3\\
Test&289&1:3\\
\bottomrule
\end{tabular}
\end{table}

For the first sub-task, the input is the problem description and the output is the set of entities that correspond to the components of the problem (Figure \ref{fig:subtask-1}). The entities are labelled according to predefined entity types. The labels were provided in Spacy format and in BIO tagging format.

For the second sub-task, the inputs are the problem description, its set of problem entities, and the order mapping of variable mentions. The ground-truth label annotations consist of the objective declaration and the constraints declarations as shown in Figure \ref{fig:subtask-2}. The output of the semantic parser is the meaning representation of those declarations. As shown in Figure \ref{fig:subtask-2}, the meaning representation should be converted to a canonical form for evaluation. Participants were encouraged to either design their own meaning representation or use the representation and conversion scripts from our pilot study. 

\paragraph{Dataset creation.} A team of 20 AI engineers and OR experts spent three months to create our preliminary LPWP dataset. This team used the Prodigy tool \citep{Prodigy} to manually create and annotate this preliminary dataset containing 600 problems. Within the team, five were tasked with verifying that each problem adhered to specific guidelines to ensure diversity in problem types and language patterns. Throughout the process of creating the remaining 501 samples, suggested annotations were generated using a preliminary NER model trained on the preliminary dataset.  For the second sub-task, we created a custom Prodigy recipe and a Python script to efficiently annotate the ground-truth declarations of the objective and constraints. All of the new problems and annotations for both sub-tasks were verified and corrected by at least two experts. \cite{EMNLP} describes in more details the data creation process (e.g., exclusion criteria, inter-annotator agreement, correction process, average duration of each step, etc.).

Note that we did not use existing datasets from third parties and have released the dataset\footnote{All data are available at: \href{https://github.com/nl4opt/nl4opt-competition}{\textcolor{blue}{https://github.com/nl4opt/nl4opt-competition}}} under the MIT License to benefit the research community.

\section{Baseline Models}

Participants had access to the code base from our pilot study that is described in more details in \cite{EMNLP}. Most participants built upon this by implementing their own methods on the provided code base.

\subsection{Sub-task 1}
The starter kit for sub-task 1 can be found in the \textsc{NL4Opt} repository\footnote{Sub-task 1 baseline is available at: \href{https://github.com/nl4opt/nl4opt-subtask1-baseline}{\textcolor{blue}{https://github.com/nl4opt/nl4opt-subtask1-baseline}}}. The baseline model, \textsc{XLM-RoBERTa-base} (\textsc{XLM-R-base}) \citep{XLM-R}, was trained and fine-tuned by minimizing the log-likelihood loss. As part of the pilot study, we reported\footnote{Stratified performance: \href{https://github.com/nl4opt/nl4opt-subtask1-baseline/tree/main/baseline\#results}{\textcolor{blue}{https://github.com/nl4opt/nl4opt-subtask1-baseline/tree/main/baseline\#results}}} the baseline model's performance on the test set when evaluated on the source domain, target domain, and entire test set for all entity types (i.e., constraint direction, limit, etc.). Based on this preliminary analysis, the objective name was the most difficult to identify potentially due to its ambiguity. We expect the greatest improvements would arise from methods that are capable of accurately recognizing the objective names and their spans. \textbf{Evaluation:} \textit{This baseline achieved an F1 score of \textbf{0.906} on the test split.}

\subsection{Sub-task 2}
The starter kit for sub-task 2 can be found in the \textsc{NL4Opt} repository\footnote{\href{https://github.com/nl4opt/nl4opt-subtask2-baseline}{Sub-task 2 baseline is available at: \textcolor{blue}{ https://github.com/nl4opt/nl4opt-subtask2-baseline}}}. The starter kit for sub-task 2 contains code to parse the XML-like intermediate representations and annotated examples into our Problem Formulation dataclass and code to score the submission. Additional information regarding the canonical representation, parsing, and scoring can be found in this \href{https://github.com/nl4opt/nl4opt-subtask2-baseline/blob/main/notebooks/demo.ipynb}{\textcolor{blue}{notebook}}. For the generation sub-task, the baseline model is a \textsc{BART} encoder-decoder \citep{BART} that leverages a prompt-guided generation and a copy mechanism to generate a meaning representation of the optimization formulation. \textbf{Evaluation:} \textit{This baseline achieved an accuracy of \textbf{0.610} on the test set. }

\section{Solutions}
\begin{table}[h]
  \begin{minipage}{.5\linewidth}
    \centering
    \caption{\label{tab:ST1-Results}Sub-task 1 winning results.}
    \small
    \begin{tabular}{ c l c }
    \\
      \toprule
      Rank & Team & F1 score \\
      \midrule
      1 & Infrrd AI Lab & 0.939 \\
      2 & mcmc & 0.933 \\
      3 & PingAn-zhiniao & 0.932 \\
      4 & Long & 0.931 \\
      5 & VTCC-NLP & 0.929 \\
      \hdashline
      - & Baseline & 0.906 \\
      \bottomrule
    \end{tabular}
  \end{minipage}%
  \begin{minipage}{.5\linewidth}
    \centering
    \caption{\label{tab:ST2-Results}Sub-task 2 winning results.}
    \small
    \begin{tabular}{ c l c }
    \\
      \toprule
      Rank & Team & F1 score \\
      \midrule
      1 & UIUC-NLP & 0.899 \\
      2 & Sjang & 0.878 \\
      3 & Long & 0.867 \\
      4 & PingAn-zhiniao & 0.866 \\
      5 & Infrrd AI Lab & 0.780 \\
      \hdashline
      - & Baseline & 0.610 \\
      \bottomrule
    \end{tabular}
  \end{minipage}
\end{table}
\subsection{Sub-task 1}
\subsubsection{First Place: Team Infrrd AI Lab}
\textsc{Team Infrrd AI Lab} (JiangLong He, Mamatha N., Shiv Vignesh, Deepak Kumar, Akshay Uppal) leveraged \textbf{ensemble learning} with \textbf{augmentation} to achieve an F1 score of \textbf{0.939} on the test set. Their base model consists of text embedding, BiLSTM, and CRF layers. Through ablations studies, they found that the \textit{majority voting} of an ensemble of 5 different models that were designed in a combination of \textsc{XLM-R-base} and \textsc{RoBERTa-base} transformers for text embeddings, BiLSTM layers, and CRF layers performed the best on the test set. They also implemented 4 types of data augmentation techniques during training. Namely, label-wise token replacement, synonym replacement, mention replacement, and shuffle within segments. For more details, refer to \citep{Infrrd}.

\subsubsection{Second Place: Team mcmc}
\textsc{Team mcmc} (Kangxu Wang, Ze Chen, Jiewen Zheng) trained models for \textbf{ensemble learning} with \textbf{adversarial attacks} to achieve an F1 score of \textbf{0.933} on the test set. They found that implementing adversarial attack using the FGM proposed by \cite{Adversarial} on the \textsc{DeBERTa-large} transformer \citep{DeBERTa-v3} with a CRF layer performed the best on the development set. They trained 9 variations of this model using different random initializations to form their ensemble and leveraged \textit{majority voting} for the final prediction. For more details, refer to \citep{mcmc}.

\subsubsection{Third Place: Team PingAn-zhiniao}
\textsc{Team PingAn-zhiniao} (Qi Zeng, Xiuyuan Yang, Yixiu Wang, Chang Shu) augmented the fine-tuning process of the \textsc{XLM-R-large} transformer by implementing a \textbf{global pointer decoder} followed by a \textbf{multi-head decoder} to achieve an F1 score of \textbf{0.932} on the test set. This was the only sub-task 1 winning submission that did not use ensemble learning. Initially, they fine-tuned using the \textsc{XLM-R-large} encoder to produce the embeddings which was fed into both the global pointer decoder and multi-head decoder. Upon reaching an F1 score of 0.9 on the development set, the global pointer decoder was removed while the encoder with multi-head decoder model continued training.

\subsubsection{Fourth Place: Team Long}
\textsc{Team Long} (Yuting Ning, Jiayu Liu, Longhu Qin, Tong Xiao, Shangzi Xue, Zhenya Huang, Qi Liu, Enhong Chen, Jinze Wu) leveraged \textbf{ensemble learning}, \textbf{adversarial training}, and some \textbf{post-processing} techniques to achieve an F1 score of \textbf{0.931} on the test set. They used \textsc{XLM-R} as the base model and leverage projected gradient descent method \citep{PGD} and FGM for adversarial training. Augmentations included variables swapping, synonym replacement in objective names, and randomizing of numbers. They also implemented some quick-check rules to enforce consistency in tagging entity spans. Four models (XLM-R-base and XLM-R-large) were optimized for specific entity types and the final prediction was obtained through an emsemble learning framework. For more details, refer to \citep{Long} or their code\footnote{\label{long}\href{https://github.com/bigdata-ustc/nl4opt}{Team Long code: \textcolor{blue}{https://github.com/bigdata-ustc/nl4opt}}}.

\subsubsection{Fifth Place: Team VTCC-NLP}
\textsc{Team VTCC-NLP} (Xuan-Dung Doan) proposed \textbf{ensemble learning} to achieve an F1 score of \textbf{0.929} on the test set. They also explored the use of ELMo embedding \citep{ELMO} and GCN models \citep{GCN} and found that both improved the performance of the baseline model accuracy, but negatively impacted the performance when included for ensemble learning. The final ensemble consisted of XLMR, DeBERTaV3, and BART. For more details, refer to \citep{VTCC}.

\subsection{Sub-task 2}
\subsubsection{First Place: Team UIUC-NLP}
\textsc{Team UIUC-NLP} (Neeraj Gangwar, Nickvash Kani) \textbf{tagged the input} and implemented a \textbf{``decode all-at-once" strategy} to achieve an accuracy of \textbf{0.899} on the reserved test set. They used the \textsc{BART-large} encoder-decoder model and enriched the input by surrounding entities with XML-like tagging. Through ablation studies, they found the best performance when combining this input tagging strategy with generating all objective and constraint declarations at once. This team also reports higher sensitivity to hyperparameters and initial seeds when using the large version of \textsc{BART} compared to the base version. For more details, refer to \citep{UIUC} or their code\footnote{\href{https://github.com/mlpgroup/nl4opt-eq-generation}{Team UIUC code: \textcolor{blue}{https://github.com/mlpgroup/nl4opt-eq-generation}}}.

\subsubsection{Second Place: Team Sjang}
\textsc{Team Sjang} (Sanghwan Jang) used a \textbf{scaling hyperparameter} to introduce \textbf{entity tag embeddings} and they implement simple \textbf{data augmentation} to achieve an accuracy of \textbf{0.878} on the reserved test set. Compared to the baseline, they report a 16\% increase in accuracy by implementing the \textsc{BART-large} model, a further 10\% improvement by scaling the tag embedding, and another 1.5\% through simple augmentations to the constraints by reversing the constraint direction. For more details, refer to \citep{Sjang} or their code\footnote{\href{https://github.com/jsh710101/nl4opt}{Team Sjang code: \textcolor{blue}{https://github.com/jsh710101/nl4opt}}}.

\subsubsection{Third Place: Team Long}
\textsc{Team Long} \textbf{redesigned the prompt}, implemented \textbf{data augmentation}, and leveraged \textbf{adversarial training} to achieve an accuracy of \textbf{0.867} on the reserved test set. They used the baseline \textsc{BART-base} with copy mechanism as the generator and leverage adversarial training during fine-tuning by using FGM. They enhance the entities by inserting XML-like tags, and alter the location of constraint and objective direction entities to where they occur in the original input description. For more details, refer to \citep{Long} or their code\textsuperscript{\ref{long}}.

\subsubsection{Fourth Place: Team PingAn-zhiniao}
\textsc{Team PingAn-zhiniao} primarily leveraged \textbf{data preprocessing} and \textbf{hyperparameter tuning} to achieve an accuracy of \textbf{0.866} on the reserved test set. Data preprocessing included wrapping entity types with tags and they report that the most improvement was brought when the bert\_dropout hyperparameter was set to 0.5.

\subsubsection{Fifth Place: Team Infrrd AI Lab}
\textsc{Team Infrrd AI Lab} \textbf{preprocessed} the input and utilized \textbf{multitask training} to achieve an accuracy of \textbf{0.780} on the reserved test set. They used the text-to-text transfer transformer (T5) \citep{T5} and processed the input by wrapping entities with the markup of entity types. They also reported an increase in performance when they separated each sample into multiple samples, each corresponding to one declaration. Multitask learning was leveraged to train the model to generate text when given different prompts. For more details, refer to \citep{Infrrd}.

\subsection{Experiments with large language models}
After the competition ended, we wanted to compare the performance of black-box large language models. In particular, we conducted some experiments with ChatGPT to see how it would perform in our competition. For these experiments, we combined the two sub-tasks and directly asked ChatGPT to generate a problem formulation from a given LPWP (problem description). We evaluated the performance of ChatGPT on both the test and development datasets using the declaration-level mapping accuracy defined in Equation (\ref{equ:accuracy}). To ensure consistent output from ChatGPT, we structured our prompts as follows:

``
$<$\textit{Problem description}$>$\
\textit{Use the above problem description and write the optimization formulation of the problem. Please only give me the model with just one-line explanations for each model element. I don't need the solution. Remove all non-essential spaces. Don't simplify the expressions and don't use LaTeX code or any code in your responses. Use ``x", ``y", and ``z" as variables name.}"

For these experiments, we used the \textit{gpt-3.5-turbo} model trained on data up to September 1st, 2021. To evaluate the performance of ChatGPT, we asked OR experts to manually verify the correctness of the generated models by ChatGPT and measured the  per-declaration accuracy. ChatGPT achieved an accuracy of \textbf{0.927} on the reserved test set for this combined task. 
\section{Discussion}

\paragraph{Sub-task 1:} Four of the top 5 teams used ensemble learning to maximize the F1 score for the NER task. While this is a great technique for competitions like \textsc{NL4Opt} that only consider performance metrics, it drastically increases the complexity which makes the training and inference more computationally expensive and less transparent. When considering methods from this competition for real-world time-sensitive applications, methods such as the Student-Teacher learning framework \citep{Wang_2022} could be explored. Other successful techniques included simple augmentation and preprocessing. Some methods also included adversarial training, or training through a two-step approach (i.e., fine-tuning using a global pointer then switch to the multi-head decoder). It is also worth pointing out that many winning teams used the transformer-based language model, DeBERTa, often as part of an ensemble. These winning methods resulted in a 2.3 to 3.3\% increase in F1 scores compared to baseline with the highest F1 score of 0.939 by \textsc{Team Infrrd AI Lab} \citep{Infrrd}.

\paragraph{Sub-task 2:} The improvements from the winning teams primarily resulted from preprocessing and data augmentation. Every winning team implemented some data augmentation or alterations to the input. The top two submissions replaced BART-base with BART-large which was responsible for higher top accuracy but a higher standard deviation was also reported. This sub-task highlights the importance of the input prompt design. We will continue to explore different input representations and the impact it has on performance. We are also interested in further exploring methods of data augmentation and training methods (i.e., ensemble learning, adversarial learning, etc.). The results of these winning submissions were encouraging as we saw a 17 to 29\% increase in declaration-level accuracy from the winning submissions with the highest accuracy of 0.899 by \textsc{Team UIUC-NLP} \citep{UIUC}.

\paragraph{Comparison with large language model}
Although ChatGPT was not trained or fine-tuned on our training set, it outperformed the winning submission of sub-task 2 by 2 percentage point. The common errors made by ChatGPT, in order of frequency of occurrence, include incorrect variable coefficients in constraints, extraneous constraints, wrong constraint directions, extra variables, missing constraints, and incorrect variable coefficients in the objective. 

The datasets used in this competition have had a lower level of complexity compared to real-world problems. As a result, it remains unclear how ChatGPT would perform when faced with more realistic and challenging problem descriptions that are frequently encountered in practical scenarios. Therefore, further research is required to examine the generalizability of large language models across a more extensive range of problem descriptions with varying levels of complexity and realism. Furthermore, it is crucial to explore methods for enhancing the trustworthiness and robustness of these models to extend their usefulness in practical applications.

\section{Conclusion}

We hosted the \textsc{NL4Opt Competition} at NeurIPS 2022 to draw attention towards the potential of machine learning in augmenting the user experience of OR tools. This competition presented two engaging tasks that successfully attracted many unique solutions. The two tasks (NER and generation) combine to take a linear programming word problem, tag its relevant entities, and generate a canonical representation that can be easily converted into a format that optimization solvers can interpret. 

To summarize, many winning teams of sub-task 1 reported a significant improvement in performance when leveraging ensemble learning and various augmentation techniques. Winning teams of sub-task 2 reported the main contributor for improved performances resulted from redesigning the input prompt. These solutions improved upon the baseline (up to 3.3\% for sub-task 1 and 29\% for sub-task 2) and will be explored for their use in making commercial solvers more accessible to non-experts by accepting natural language problem descriptions. ChatGPT achieved a 2.8\% improvement over the top-performing submission for sub-task 2 without the need for intermediate entity tagging. Future research should investigate the generalizability and robustness of large language models and their potential benefits for OR applications.

In addition to the impact of providing an alternative input format to solvers, the labelled dataset from this competition has been released and may be used to evaluate methods for multi-sentence inputs, low-resource learning (eg. zero-shot/few-shot learning), and generalizability to unseen domains. We encourage and look forward to continual applications of the open-sourced dataset and the subsequent exciting new research interests that may stem from the solutions of the \textsc{NL4Opt Competition}.


\acks{A big thank you to all the participants of the competition for their interest and engagement. This competition was a success thanks to your hard work. We would also like to acknowledge the effort of the co-organizers from University of British Columbia, Ivey Business School, and Huawei Technologies Canada Co. Ltd.}


\newpage
\bibliography{anthology.bib, custom.bib}

\end{document}